\documentclass{article}
\usepackage{ijcai26}

\usepackage{times}
\usepackage{soul}
\usepackage{url}
\usepackage[hidelinks]{hyperref}
\usepackage[utf8]{inputenc}
\usepackage[small]{caption}
\usepackage{graphicx}
\usepackage{amsmath}
\usepackage{amsthm}
\usepackage{booktabs}
\usepackage{algorithm}
\usepackage{algorithmic}
\usepackage[switch]{lineno}
\usepackage{placeins} 
\usepackage{caption}
\usepackage{subcaption}
\usepackage{fontawesome5}

\usepackage[table]{xcolor} 
\usepackage{subcaption}
\definecolor{gray}{HTML}{E0E0E0}
\definecolor{blue}{HTML}{e9edf6}
\definecolor{red}{HTML}{C00000}
\definecolor{green}{HTML}{548235}
\definecolor{deeppurple}{HTML}{5B2C83}
\usepackage[breakable, skins, most]{tcolorbox}
\usepackage{tabularx}  
\usepackage{enumitem}  

\DeclareMathOperator*{\argmax}{arg\,max}

\title{Video-MSR: Benchmarking Multi-hop Spatial Reasoning Capabilities of MLLMs}

\author{
Rui Zhu\textsuperscript{1,2} \quad
Xin Shen\textsuperscript{1,3} \quad
Shuchen Wu\textsuperscript{1} \quad
Chenxi Miao\textsuperscript{1} \quad
Xin Yu\textsuperscript{3} \quad
Yang Li\textsuperscript{1}\thanks{Corresponding author.} \quad
Weikang Li\textsuperscript{4} \quad
Deguo Xia\textsuperscript{1} \quad
Jizhou Huang\textsuperscript{1} \quad
\affiliations
\textsuperscript{1}Baidu Inc. \quad
\textsuperscript{2}Nanjing University \quad
\textsuperscript{3}The University of Queensland \quad
\textsuperscript{4}Peking University \quad
{\tt\small liyang164@baidu.com}
}

\begin{document}
\maketitle

\begin{abstract}
Spatial reasoning has emerged as a critical capability for Multimodal Large Language Models (MLLMs), drawing increasing attention and rapid advancement. 
However, existing benchmarks primarily focus on single-step perception-to-judgment tasks, leaving scenarios requiring complex visual-spatial logical chains significantly underexplored. 
To bridge this gap, we introduce \textbf{\textit{Video-MSR}}, the first benchmark specifically designed to evaluate \textbf{\textit{Multi-hop Spatial Reasoning (MSR)}} in dynamic video scenarios. 
Video-MSR systematically probes MSR capabilities through four distinct tasks: \textit{Constrained Localization, Chain-based Reference Retrieval, Route Planning, and Counterfactual Physical Deduction}. 
Our benchmark comprises 3,052 high-quality video instances with 4,993 question-answer pairs, constructed via a scalable, visually-grounded pipeline combining advanced model generation with rigorous human verification. 
Through a comprehensive evaluation of 20 state-of-the-art MLLMs, we uncover significant limitations, revealing that while models demonstrate proficiency in surface-level perception, they exhibit distinct performance drops in MSR tasks, frequently suffering from spatial disorientation and hallucination during multi-step deductions. 
To mitigate these shortcomings and empower models with stronger MSR capabilities, we further curate \textbf{\textit{MSR-9K}}, a specialized instruction-tuning dataset, and fine-tune Qwen-VL, achieving a +7.82\% absolute improvement on Video-MSR. 
Our results underscore the efficacy of multi-hop spatial instruction data and establish Video-MSR as a vital foundation for future research. 
The code and data will be available at~\href{https://github.com/ruiz-nju/Video-MSR}{\faGithub~\textcolor{deeppurple}{Video-MSR}}.
\end{abstract}

\begin{figure*}[ht]
    \begin{center}
        \includegraphics[width=1\linewidth]{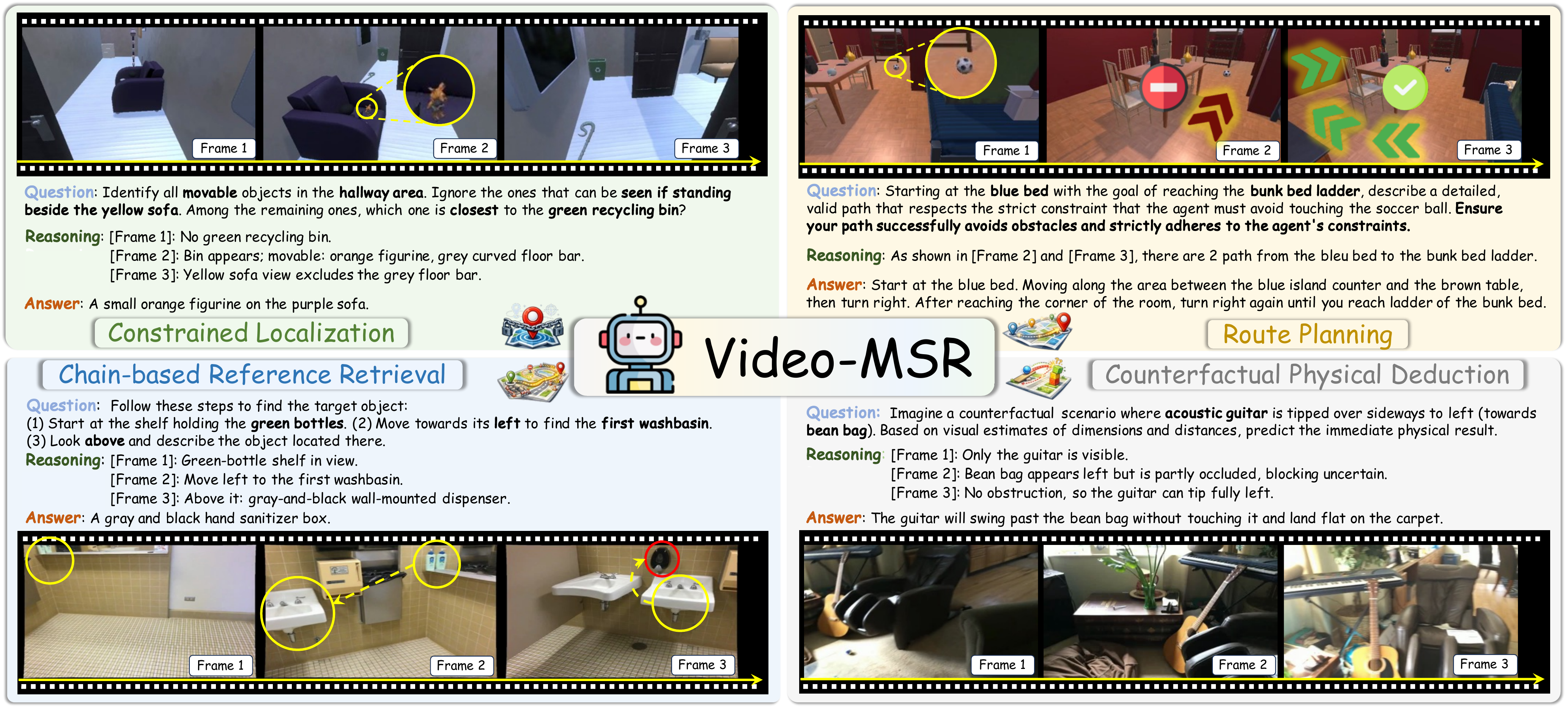}
    \end{center}
    \caption{\textbf{Video-MSR benchmark overview.} To systematically estimated the spatial reasoning capabilities of MLLMs, we introduce four distinct tasks: Constrained Localization, Chain-based Reference Retrieval, Route Planning, and Counterfactual Physical Deduction.}
    \label{fig:teaser}
\end{figure*}

\section{Introduction}

\begin{quote}
\raggedright %
{\em Spatial thinking, rooted in perception of space and action in it, is the foundation for all thought.}\\
\hfill-- {\scriptsize Barbara Tversky, \emph{Mind in Motion: How Action Shapes Thought}} %
\end{quote}

Spatial thinking serves as a cornerstone of human intelligence, enabling us not only to navigate the physical world but also to structure abstract concepts and solve complex logical problems~\cite{mindInMotion}.
From inferring 3D structures via 2D retinal projections to planning multi-step routes in dynamic environments, this capability forms the foundation of our interaction with reality. 
In the realm of artificial intelligence, Multimodal Large Language Models (MLLMs) have demonstrated remarkable proficiency in visual perception and semantic understanding \cite{Qwen-VL,internvl,llava,gpt4o,gemini25}. 
However, true visual intelligence extends beyond merely recognizing objects; it demands the ability to understand and manipulate 2D and 3D spatial relationships to derive logical conclusions~\cite{kamath2023whatsup,spatialVLM,vsibench,why_spatial_hard}. 
This spatial reasoning capability is becoming increasingly pivotal, serving as a prerequisite for high-level reasoning and interaction in downstream tasks, such as embodied planning and manipulation~\cite{CoT-VLA}. 
Therefore, systematically exploring the spatial reasoning capacity of MLLMs is essential to identify the critical gap between superficial visual perception and deep logical reasoning, thereby pushing the boundaries of current multimodal intelligence.

To evaluate the spatial reasoning proficiency of MLLMs, existing works have primarily investigated their performance across fundamental tasks, such as object counting, perspective transformation, relative positional awareness, and depth estimation~\cite{vsibench,cambrian-1-cvbench,vsp,llava-st,cambrian-s}. 
While necessitating accurate visual perception, these tasks remain largely confined to a simplistic ``perception-to-judgment'' paradigm, resolvable through \textit{single-step inference} based on explicit cues. 
However, advanced spatial intelligence demands a higher order of cognition--specifically, \textbf{\textit{Multi-hop Spatial Reasoning (MSR)}}--which involves maintaining intermediate reasoning states, synthesizing visual cues into coherent mental models, and strictly adhering to spatial constraints across long-horizon logical chains. 
The capability of MLLMs to perform such rigorous MSR reasoning remains underexplored. Consequently, this underscores an urgent need to address two critical questions: \textit{(i) To what extent do MLLMs possess the capacity for MSR? and (ii) What are the primary bottlenecks limiting their performance in MSR?}

To bridge this gap, we curate \textbf{\textit{Video-MSR}}, the first benchmark specifically designed to evaluate the MSR performance of MLLMs in videos. 
As shown in Figure \ref{fig:teaser}, Video-MSR encompasses four distinct question types: \textit{Constrained Localization}, \textit{Chain-based Reference Retrieval}, \textit{Route Planning}, and \textit{Counterfactual Physical Deduction}. 
The first three tasks rigorously test the model's ability to navigate and localize objects under strict spatial constraints and multi-step reference chains, while the counterfactual task challenges the model to predict physical interactions and geometric conflicts in hypothetical scenarios. 
Our benchmark is sourced from 3,052 diverse video instances collected from six public datasets \cite{scannet-dataset,scannetpp-dataset,Arkitscenes-dataset,adt-dataset,procthor-dataset,s3dis-dataset}. 
To ensure high quality, we designed a visually-grounded annotation pipeline, leveraging state-of-the-art MLLMs (\emph{e.g.}, Gemini-3-Pro) for automated generation followed by rigorous human verification, resulting in 4,993 high-quality question-answer pairs. 
Each pair demands complex multi-hop spatial reasoning strictly grounded in video content. 

We further evaluate 20 state-of-the-art MLLMs, ranging from lightweight open-source models to proprietary systems such as GPT-4o~\cite{gpt4o} and Gemini~\cite{gemini25}. 
While many models perform competitively on surface-level perception tasks, they exhibit distinct performance drops on MSR. 
Specifically, we observe that models often fail to maintain logical consistency, suffering from severe spatial disorientation and hallucination during multi-step reasoning. 
These failures reveal a lack of deep internalization of combined spatial perception and logic in current architectures. 
To mitigate these shortcomings and empower models with stronger MSR capabilities, we futher curate \textbf{MSR-9K}, a specialized instruction-tuning dataset. 
By fine-tuning Qwen-VL~\cite{Qwen2.5-VL,Qwen3-VL} on MSR-9K, we achieve a 7.82\% absolute improvement on Video-MSR, validating the efficacy of our proposed data construction paradigm.

In summary, our contributions are summarized as follows: 
\begin{itemize}
    \item We define the task of \textbf{Multi-hop Spatial Reasoning (MSR)}, requiring models to synthesize spatial cues and construct multi-step reasoning chains to resolve complex spatial dynamics.
    \item We introduce \textbf{Video-MSR}, the first benchmark dedicated to MSR evaluation. It spans four distinct reasoning tasks and comprises 3,052 video instances with 4,993 QA pairs, constructed via a scalable, visually-grounded generation pipeline. 
    \item We conduct a comprehensive evaluation of 20 representative MLLMs, uncovering significant gaps between surface-level perception and deep spatial logic, while providing actionable insights into the spatial disorientation bottlenecks of current systems. 
    \item We curate \textbf{MSR-9K}, a high-quality instruction-tuning dataset designed to enhance multi-hop spatial reasoning. Our experiments demonstrate that fine-tuning on MSR-9K yields substantial performance gains (\emph{e.g.}, \textbf{7.82\%} increment on Video-MSR), establishing a strong baseline for future research. 
\end{itemize}

\begin{figure*}[ht]
    \begin{center}
        \includegraphics[width=\linewidth]{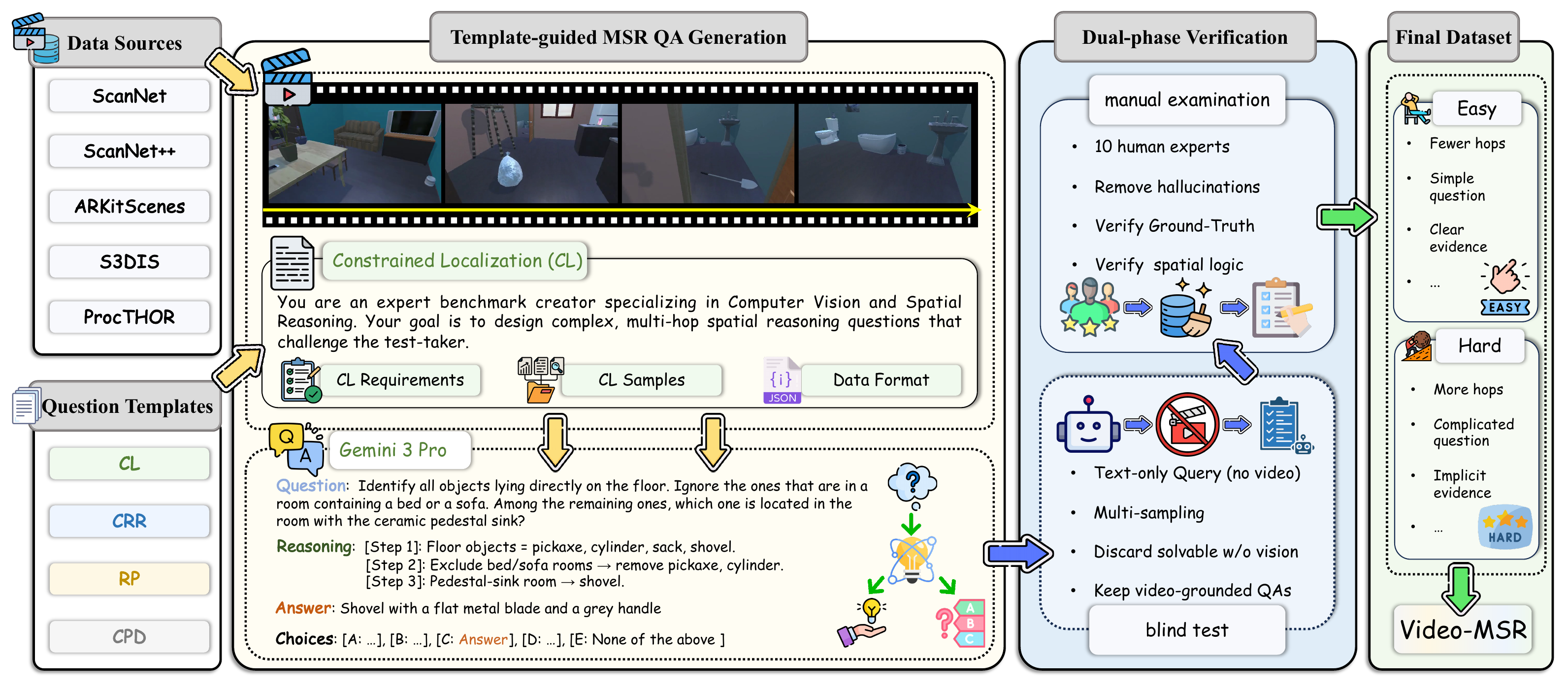}
    \end{center}
    \caption{\textbf{Overview of the data construction pipeline.} We integrate automated generation with dual-phase verification to produce high-quality MSR data. Starting with high-fidelity video collection, we employ Gemini-3.0-Pro as the core engine to generate template-guided QA pairs. A subsequent dual-phase filtration process, including a text-only blind test and manual expert verification, is applied to eliminate biases and hallucinations, ensuring the reliability of multi-hop spatial reasoning chains.}
    \label{fig:annotation_pipeline}
\end{figure*}

\section{Multi-hop Spatial Reasoning}
In this section, we introduce Multi-hop Spatial Reasoning (MSR), which requires models to synthesize spatial cues and construct multi-step reasoning chains to resolve complex spatial problems.

\subsection{Problem Definition}

Formally, let $\mathcal{V}$ denote the visual input (\emph{i.e.}, a video clip) and $\mathcal{Q}$ represent the natural language query. Unlike conventional single-hop spatial tasks where the answer $\mathcal{A}$ can be derived directly via a mapping $P(\mathcal{A} | \mathcal{V}, \mathcal{Q})$, MSR necessitates the construction of an explicit spatial reasoning chain to bridge the gap between perception and the final answer.

We model this process as sequential transitions of spatial states, denoted as $\mathcal{S} = \{s_0, s_1, \dots, s_T\}$, where $T$ represents the number of reasoning hops. Here, $s_0$ represents the initial anchor derived from $\mathcal{Q}$, and each subsequent state $s_t$ corresponds to an intermediate spatial entity, region, or physical status identified during the reasoning process. 

The generation of the reasoning chain follows a \textit{Markovian property}, where the identification of the current spatial state $s_t$ depends on the immediately preceding state $s_{t-1}$, the specific spatial constraint $c_t$ extracted from $\mathcal{Q}$, and the visual evidence in $\mathcal{V}$. This transition can be formulated as:
\begin{equation}
    P(s_t | s_{t-1}, \dots, s_0, \mathcal{V}, \mathcal{Q}) \approx P(s_t | s_{t-1}, \mathcal{V}, c_t),
\end{equation}
where $P(s_t | s_{t-1}, \mathcal{V}, c_t)$ represents the probability of locating the $t$-th spatial node given the previous node and the current spatial constraint.

Consequently, the overall objective is to maximize the joint probability of the reasoning chain and the final answer ($\mathcal{A}^*$):
\begin{equation}
    \mathcal{A}^* = \argmax_{\mathcal{A}} P(\mathcal{A} | s_T, \mathcal{V}) \prod_{t=1}^{T} P(s_t | s_{t-1}, \mathcal{V}, c_t).
\end{equation}
In this formulation, success demands the recursive extraction of spatial features to forge a coherent reasoning chain, guaranteeing that the final output accurately reflects a precise internalization and application of visual-spatial evidence.

\begin{figure*}[t]
\centering
\newcommand{\contentht}{3.5cm} 
\begin{minipage}[t]{0.37\textwidth}
\vspace{0pt}
\centering
\parbox[c][\contentht][c]{\linewidth}{%
  \centering
  \resizebox{\linewidth}{!}{%
  \begin{tabular}{lrrr}
      \toprule
      \textbf{Data source} & \textbf{\# Videos} & \textbf{\# QA Pairs} & \textbf{\# Easy / Hard} \\
      \midrule
      ADT & 48 & 77 & 51 / 26 \\
      ARKitScenes & 129 & 207 & 93 / 114 \\
      ProcTHOR & 2,358 & 3,844 & 1,640 / 2,204 \\
      S3DIS & 179 & 299 & 135 / 164 \\
      ScanNet & 292 & 487 & 204 / 283 \\
      ScanNet++ & 46 & 79 & 29 / 50 \\
      \midrule
      \textbf{Total} & \textbf{3,052} & \textbf{4,993} & \textbf{2,152 / 2,841} \\
      \bottomrule
  \end{tabular}}
}
\captionof{table}{Data statistics of Video-MSR. The table details the number of video, QA pairs, and the difficulty distribution across different source datasets.}
\label{tab:statistics}
\end{minipage}\hfill
\begin{minipage}[t]{0.22\textwidth}
\vspace{0pt}
\centering
\parbox[c][\contentht][c]{\linewidth}{%
  \centering
  \includegraphics[width=\linewidth]{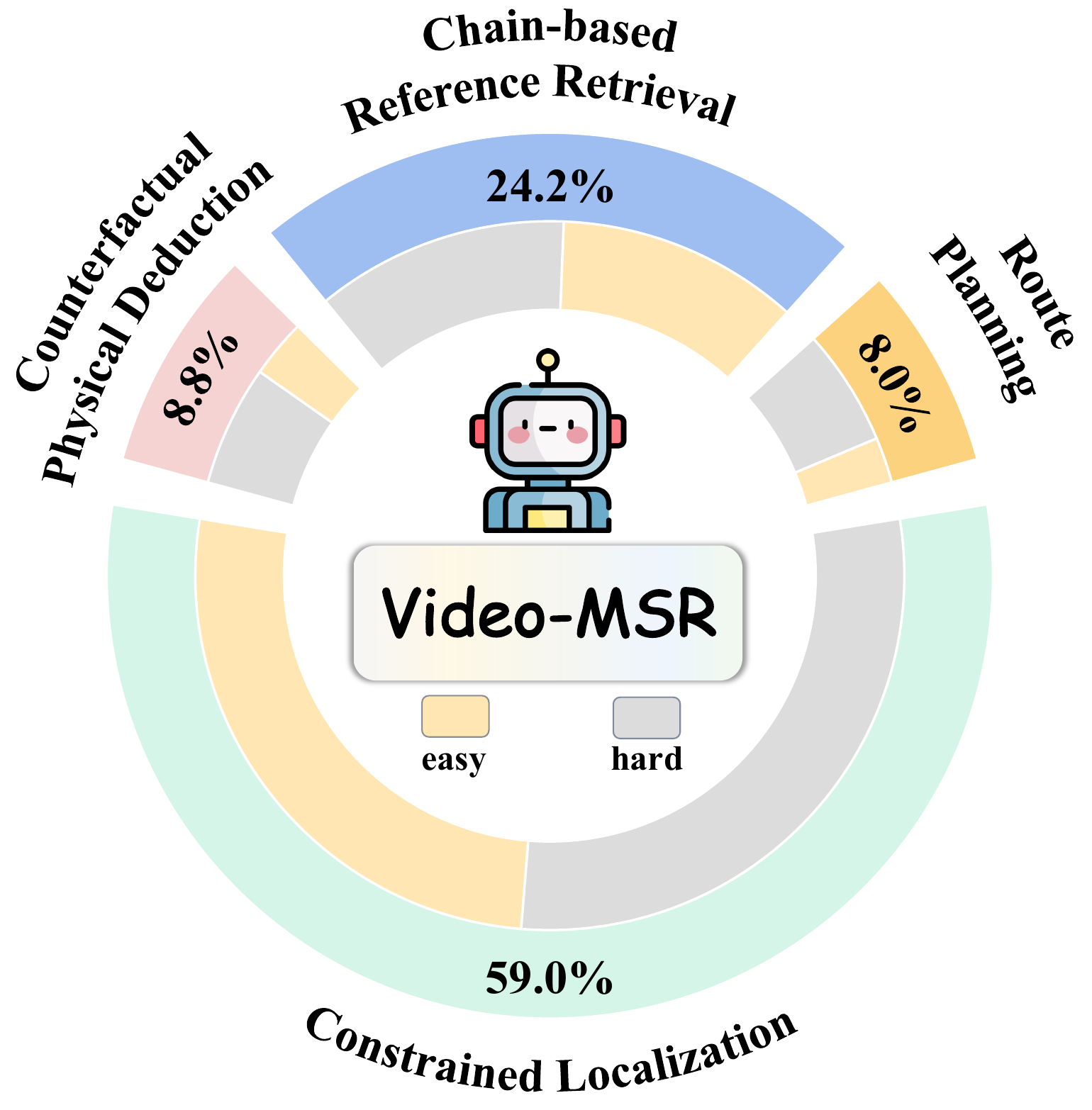}%
}
\captionof{figure}{Distribution of QA pairs across four tasks.}
\label{fig:task_dist}
\end{minipage}\hfill
\begin{minipage}[t]{0.37\textwidth}
\vspace{0pt}
\centering
\parbox[c][\contentht][c]{\linewidth}{%
  \centering
  \includegraphics[width=\linewidth]{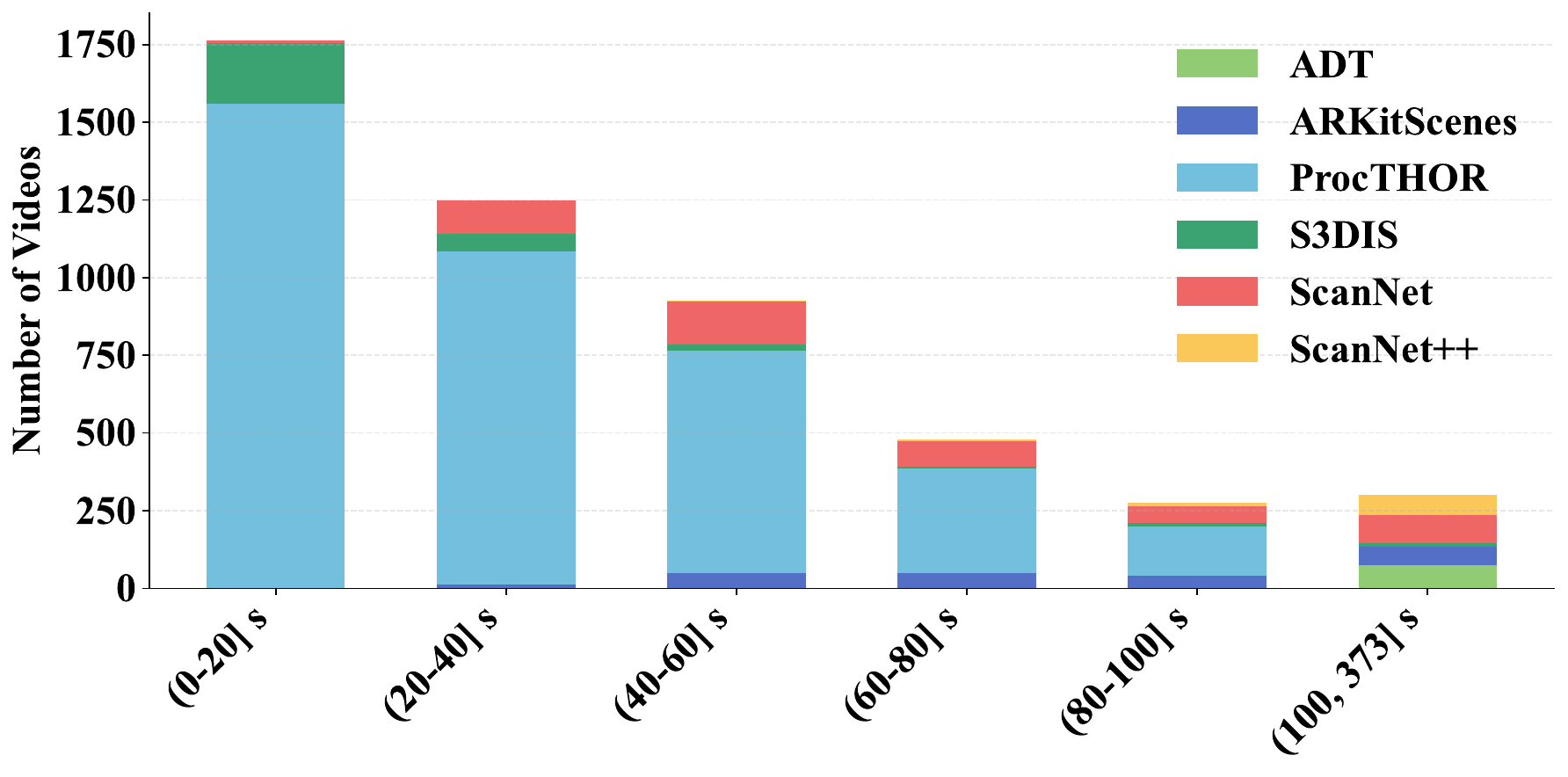}%
}
\captionof{figure}{Distribution of video lengths across different source datasets in Video-MSR.}
\label{fig:len_dist}
\end{minipage}

\end{figure*}

\begin{table*}[h]
    \centering
    \resizebox{\linewidth}{!}{
        \begin{tabular}{l|ccc|ccc|ccc|ccc|ccc}
            \toprule
            & \multicolumn{3}{c|}{\textbf{Overall}} & \multicolumn{3}{c|}{\textbf{CL}} & \multicolumn{3}{c|}{\textbf{CRR}} & \multicolumn{3}{c|}{\textbf{RP}} & \multicolumn{3}{c}{\textbf{CPD}}  \\
            \cmidrule(lr){2-4} \cmidrule(lr){5-7} \cmidrule(lr){8-10} \cmidrule(lr){11-13}  \cmidrule(lr){14-16}
            \textbf{Model} & Avg. & Easy & Hard & Avg. & Easy & Hard & Avg. & Easy & Hard & Avg. & Easy & Hard & Avg. & Easy & Hard \\
            \midrule
            \rowcolor[HTML]{e9edf6}
            \multicolumn{16}{c}{\textit{MLLMs:}  \textless \ 7B} \\
            \midrule
            Qwen3-VL-4B-Instruct & 40.00 & \textbf{55.95} & 27.91 & 36.85 & \textbf{51.78} & 23.25 & 52.24 & \underline{67.17} & 37.93 & 32.24 & 49.11 & 25.61 & 34.47 & 55.81 & 32.15 \\
            Qwen2.5-VL-3B-Instruct & 30.02 & 35.73 & 25.70 & 33.32 & 38.62 & 28.50 & 27.98 & 34.01 & 22.20 & 4.79 & 5.36 & 4.56 & 36.30 & 44.19 & 35.44 \\
            InternVL3.5-4B-Instruct & 24.11 & 28.90 & 20.49 & 28.61 & 33.57 & 24.09 & 19.54 & 21.49 & 17.67 & 4.53 & 7.14 & 3.51 & 24.20 & 34.88 & 23.04 \\
            InternVL3.5-2B-Instruct & 22.06 & 24.59 & 20.13 & 26.71 & 29.37 & 24.29 & 14.58 & 16.27 & 12.97 & 1.01 & 1.79 & 0.70 & 30.37 & 41.86 & 29.11 \\
            InternVL3-2B-Instruct & 23.45 & 26.44 & 21.19 & 29.73 & 33.50 & 26.30 & 13.16 & 13.37 & 12.97 & 0.50 & 0.00 & 0.70 & 30.37 & 44.19 & 28.86 \\
            \midrule
            \rowcolor[HTML]{e9edf6}
            \multicolumn{16}{c}{\textit{MLLMs:} 7B - 10B} \\
            \midrule
            Qwen3-VL-8B-Instruct & \textbf{43.92} & \underline{52.70} & \textbf{37.28} & \textbf{38.44} & \underline{47.72} & \underline{29.99} & 55.79 & 64.13 & 47.81 & 54.66 & 55.36 & \underline{54.39} & \textbf{38.36} & \textbf{51.16} & \textbf{36.96} \\
            Qwen2.5-VL-7B-Instruct & 33.23 & 40.52 & 27.70 & 32.54 & 39.33 & 26.36 & 41.31 & 48.73 & 34.20 & 10.83 & 10.71 & 10.88 & 35.84 & 44.19 & 34.94 \\
            InternVL3.5-8B-Instruct & 27.88 & 33.13 & 23.90 & 31.80 & 37.48 & 26.62 & 25.00  & 27.24 & 22.85 & 8.06 & 8.04 & 8.07 & 27.40 & 37.21 & 26.33 \\
            InternVL3-8B-Instruct & 26.84 & 31.51 & 23.30 & 30.98 & 36.13 & 26.30 & 21.85 & 25.21 & 18.64 & 3.78 & 4.46 & 3.51 & 33.56 & 37.21 & 33.16 \\
            LLaVA-NeXT-Video-7B & 19.63 & 20.91 & 18.66 & 28.07 & 29.37 & 26.88 & 3.73 & 4.57 & 2.92 & 0.00 & 0.00 & 0.00 & 24.43 & 23.26 & 24.56 \\
            \midrule
            \rowcolor[HTML]{e9edf6}
            \multicolumn{16}{c}{\textit{MLLMs:} 10B - 30B} \\
            \midrule
            InternVL3.5-14B-Instruct & 22.81 & 26.81 & 19.78 & 24.07 & 28.59 & 19.95 & 21.11 & 23.86 & 18.48 & 12.85 & 16.96 & 11.23 & 28.08 & 34.88 & 27.34 \\
            InternVL3-14B-Instruct & 28.16 & 34.29 & 23.51 & 32.20 & 38.41 & 26.55 & 25.00 & 29.78 & 20.42 & 7.05 & 8.93 & 6.32 & 28.77 & 27.91 & 28.86 \\
            \midrule
            \rowcolor[HTML]{e9edf6}
            \multicolumn{16}{c}{\textit{MLLMs:} 30B - 70B} \\
            \midrule
            Qwen3-VL-32B-Instruct & \underline{42.95} & 50.46 & \underline{37.26} & 36.07 & 43.88 & 28.95 & \underline{57.37} & 63.11 & \underline{51.86} & \underline{57.93} & \textbf{63.11} & 51.86 & 30.14 & 46.51 & 28.35 \\
            Qwen2.5-VL-32B-Instruct & 37.12 & 42.82 & 32.81 & 35.90 & 42.39 & \underline{29.99} & 36.79 & 41.36 & 32.41 & 47.86 & 53.57 & 45.61 & \underline{36.53} & \underline{48.84} & \underline{35.19} \\
            InternVL3.5-38B-Instruct & 26.24 & 30.86 & 22.74 & 24.37 & 27.67 & 21.37 & 35.43 & 39.09 & 31.93 & 17.38 & 24.11 & 14.74 & 21.46 & 39.53 & 19.49 \\
            InternVL3-38B-Instruct & 31.15 & 36.73 & 26.93 & 35.25 & 41.04 & \underline{29.99} & 27.09 & 30.51 & 23.82 & 9.57 & 11.61 & 8.77 & 34.25 & 46.51 & 32.91 \\
            \midrule
            \rowcolor[HTML]{e9edf6}
            \multicolumn{16}{c}{\textit{MLLMs:} \textgreater \ 70B} \\
            \midrule
            Qwen2.5-VL-72B-Instruct & 37.28 & 44.12 & 32.10 & 33.86 & 41.04 & 27.33 & 46.48 & 51.36 & 41.82 & 37.03 & 41.96 & 35.09 & 35.16 & \textbf{51.16} & 33.42 \\
            InternVL3-78B-Instruct & 32.57 & 39.17 & 27.57 & 35.36 & 42.53 & 28.82 & 31.90 & 34.69 & 29.22 & 13.10 & 16.07 & 11.93 & 33.33 & \textbf{51.16} & 31.39 \\
            \midrule
            \rowcolor[HTML]{e9edf6}
            \multicolumn{16}{c}{\textit{MLLMs:} Proprietary} \\
            \midrule
            Gemini-2.5-Flash & 37.02 & 39.90 & 34.85 & 23.58 & 25.37 & 21.96 & \textbf{66.33} & \textbf{70.68} & \textbf{62.18} & \textbf{61.21} & \underline{60.71} & \textbf{61.40} & 24.77 & 37.21 & 23.41 \\
            GPT-4o & 41.87 & 49.02 & 36.44 & \underline{38.07} & 44.95 & \textbf{31.80} & 55.76 & 60.07 & 51.62 & 45.84 & 46.43 & 45.61 & 25.57 & 37.21 & 24.30 \\
            \bottomrule
        \end{tabular}
    }
    \caption{\textbf{Main evaluation results on the Video-MSR benchmark.} We report performance across four distinct spatial reasoning tasks: Constrained Localization (CL), Chain-based Reference Retrieval (CRR), Route Planning (RP), and Counterfactual Physical Deduction (CPD). Models are grouped by parameter. The best and second-best results within each group are highlighted in \textbf{bold} and \underline{underlined}, respectively.}
    \label{tab:main_results}
\end{table*}

\section{Video-MSR}
To facilitate a systematic evaluation of MLLMs on MSR problems, we present \textit{Video-MSR}, the first comprehensive benchmark tailored to probe the boundaries of multi-hop spatial reasoning capabilities in dynamic video scenarios.

\subsection{Taxonomy}

Video-MSR encompasses four distinct task categories: \textit{Constrained Localization}, \textit{Chain-based Reference Retrieval}, \textit{Route Planning}, and \textit{Counterfactual Physical Deduction}. Each task is meticulously designed to evaluate a specific dimension of spatial reasoning.

\paragraph{Constrained Localization (CL).}
CL evaluates the capability of MLLMs to perform exclusionary spatial reasoning within a cluttered scene. Unlike standard referring expression comprehension that targets a unique object based on direct description, CL requires a multi-step filtering process: the model must first identify a broad set of candidate objects, then exclude specific instances that satisfy a spatial constraint, and finally pinpoint the target among the remaining candidates based on a secondary condition. This tests the model's ability to handle logical negation and set-based spatial operations.

\paragraph{Chain-based Reference Retrieval (CRR).}
CRR is designed to probe the model's ability to handle sequential spatial dependencies. In this task, the target object is not described by its intrinsic features but by its spatial relationship to a sequence of landmarks. The model must initiate the reasoning process at a distinct anchor object, trace a specific directional path, and identify intermediate nodes to eventually locate the target. This task strictly follows a Markovian property, where successful localization depends entirely on the accurate maintenance of intermediate spatial states, making it highly sensitive to error propagation and attention drift.

\paragraph{Route Planning (RP).}
Mimicking real-world embodied navigation, RP challenges the model to generate a feasible trajectory from a start point to a goal while strictly adhering to physical or semantic constraint. This requires the model to construct a coherent 3D mental map of the environment, identify traversable and non-traversable zones based on visual cues, and synthesize a continuous path that respects these spatial boundaries. It moves beyond static recognition to evaluate the model's capacity for spatial planning and obstacle avoidance in a potential action space.

\paragraph{Counterfactual Physical Deduction (CPD).}
This task assesses the model's predictive capability regarding dynamic spatial interactions. It presents a hypothetical scenario involving an object's motion and asks the model to deduce the physical consequence based on the static visual evidence. Solving CPD requires the model to mentally simulate the object's trajectory, estimate relative distances and dimensions in 3D space, and detect potential geometric interference, thereby testing the internalization of physical common sense and spatial kinematics.

\subsection{Benchmark Construction}

As illustrated in Figure \ref{fig:annotation_pipeline}, we develop a sophisticated construction pipeline to generate high-quality, video-grounded QA pairs for MSR problems. Our pipeline integrates automated generation with multi-stage verification, ensuring scalability while maintaining strict adherence to spatial logic.

\paragraph{Data Collection.} To ensure the quality and diversity of our benchmark, we meticulously curate a comprehensive collection of videos from established public datasets, including ScanNet~\cite{scannet-dataset}, ScanNet++~\cite{scannetpp-dataset}, ARKitScenes~\cite{Arkitscenes-dataset},
ADT~\cite{adt-dataset}, S3DIS~\cite{s3dis-dataset}, and ProcTHOR~\cite{procthor-dataset}. These datasets provide high-fidelity video scans that encompass a rich spectrum of environments, ranging from controlled synthetic simulations to complex real-world indoor scenes captured via handheld cameras or professional scanners. Crucially, the inherent complexity and object diversity within these scenes compel MLLMs to strictly adhere to multifaceted spatial constraints and construct rigorous spatial reasoning chains to derive correct answers, rather than relying on shallow object recognition.

\paragraph{Question-Answer Generation.} To balance generation efficiency with quality, we first manually design distinct templates for each of the four task categories. These templates serve as structural frameworks to standardize the output format, guiding the model to adhere to specific problem formulations. Subsequently, we employ Gemini-3.0-Pro as our core generation engine. By feeding both the video clips and specific task templates into the model, we prompt it to synthesize questions that strictly align with the unique logic of each task category. Crucially, the model is instructed to ground these questions in actual visual details, ensuring they necessitate complex spatial reasoning chains rather than simple visual matching. Detailed workflows and specific template designs are provided in the supplementary material.

\paragraph{Quality Control.} We implement a dual-phase verification strategy to ensure data reliability and robustness. The first phase involves an automated blind test using Gemini-3.0-Pro. Here, we input only the textual queries to the model while withholding visual data, performing multiple sampling iterations to detect potential biases. Any question consistently solvable without the corresponding video context is discarded. This rigorous filtration guarantees that the reasoning process strictly relies on the spatial information within the video, thereby preserving the video-grounded nature of the QA pairs. Following this automated step, we recruit 10 human experts to conduct a final manual examination. This step aims to eliminate potential hallucinations and verify the logical correctness of the ground truth answers.

\subsection{Data Statistics}

Following the construction pipeline, Video-MSR comprises 4,993 QA pairs grounded in 3,052 video clips, covering four distinct tasks. Detailed statistics are presented in Table \ref{tab:statistics} and Figure \ref{fig:task_dist}. We also report the video length distribution in Figure \ref{fig:len_dist}. To support a fine-grained evaluation, we stratify the tasks into \textit{easy} and \textit{hard} levels based on the required reasoning steps and video duration. Hard questions necessitate intricate reasoning chains, often involving distracting visual information within the video. Conversely, easy questions can be resolved with fewer visual cues and simpler reasoning logic.

To further empower models with stronger MSR capabilities, we curate \textbf{MSR-9K}, a specialized instruction-tuning dataset containing 8,369 MSR QA pairs. Detailed information is provided in the supplementary material.

\begin{table*}[h]
    \centering
    \resizebox{\linewidth}{!}{
        \begin{tabular}{l|ccc|ccc|ccc|ccc|ccc}
            \toprule
            & \multicolumn{3}{c|}{\textbf{Overall}} & \multicolumn{3}{c|}{\textbf{CL}} & \multicolumn{3}{c|}{\textbf{CRR}} & \multicolumn{3}{c|}{\textbf{RP}} & \multicolumn{3}{c}{\textbf{CPD}}  \\
            \cmidrule(lr){2-4} \cmidrule(lr){5-7} \cmidrule(lr){8-10} \cmidrule(lr){11-13}  \cmidrule(lr){14-16}
            \textbf{Model} & Avg. & Easy & Hard & Avg. & Easy & Hard & Avg. & Easy & Hard & Avg. & Easy & Hard & Avg. & Easy & Hard \\
            \midrule
            Cambrian-S-7B-Base & 32.91 & 39.36 & 28.02 & 37.08 & 43.24 & 31.48 & 29.14 & 35.19 & 23.34 & 6.05 & 5.36 & 6.32 & 39.50 & 58.14 & 37.47 \\
            Cambrian-S-7B & 27.82 & 32.85 & 24.01 & 30.47 & 35.28 & 26.10 & 25.91 & 30.96 & 21.07 & 3.78 & 2.68 & 4.21 & 36.99 & 58.14 & 34.68 \\
            \midrule
            Qwen2.5-VL-7B-Instruct & 33.23 & 40.52 & 27.70 & 32.54 & 39.33 & 26.36 & 41.31 & 48.73 & 34.20 & 10.83 & 10.71 & 10.88 & 35.84 & 44.19 & 34.94 \\
            \textit{+ CoT} & 31.43 & 36.68 & 27.46 & 29.86 & 35.42 & 24.81 & 38.11 & 42.54 & 33.87 & 23.17 & 21.43 & 23.86 & 31.05 & 37.21 & 30.38 \\
            \textit{+ MSR-9K-SFT (Ours)} & 41.05 & 45.31 & 37.83 & 35.97 & 39.54 & 32.71 & 45.99 & 55.18 & 37.28 & 59.45 & 62.50 & 58.25 & 44.98 & 53.49 & 44.05 \\
            \rowcolor{gray}
            $\Delta$ & \textcolor{red}{+ 7.82} & \textcolor{red}{+ 4.79} & \textcolor{red}{+ 10.13} & \textcolor{red}{+ 3.43} & \textcolor{red}{+ 0.21} & \textcolor{red}{+ 6.35} & \textcolor{red}{+ 4.68} & \textcolor{red}{+ 6.45} & \textcolor{red}{+ 3.08} & \textcolor{red}{+ 48.62} & \textcolor{red}{+ 51.79} & \textcolor{red}{+ 47.37} & \textcolor{red}{+ 9.14} & \textcolor{red}{+ 9.30} & \textcolor{red}{+ 9.11} \\
            \midrule
            Qwen3-VL-8B-Instruct & 43.92 & 52.70 & 37.28 & 38.44 & 47.72 & 29.99 & 55.79 & 64.13 & 47.81 & 54.66 & 55.36 & 54.39 & 38.36 & 51.16 & 36.96 \\
            \textit{+ CoT} & 39.38 & 45.96 & 34.39 & 34.61 & 42.03 & 27.85 & 51.49 & 56.01 & 47.16 & 45.59 & 42.86 & 46.67 & 32.42 & 44.19 & 31.14 \\
            \textit{+ MSR-9K-SFT (Ours)} & 47.02 & 55.39 & 40.68 & 41.05 & 50.97 & 31.99 & 56.49 & 64.06 & 49.27 & 66.25 & 65.18 & 66.67 & 43.61 & 53.49 & 42.53 \\
            \rowcolor{gray}
            $\Delta$ & \textcolor{red}{+ 3.10} & \textcolor{red}{+ 2.69} & \textcolor{red}{+ 3.40} & \textcolor{red}{+ 2.61} & \textcolor{red}{+ 3.25} & \textcolor{red}{+ 2.00} & \textcolor{red}{+ 0.70} & \textcolor{green}{- 0.07} & \textcolor{red}{+ 1.46} & \textcolor{red}{+ 11.59} & \textcolor{red}{+ 9.82} & \textcolor{red}{+ 12.28} & \textcolor{red}{+ 5.25} & \textcolor{red}{+ 2.33} & \textcolor{red}{+ 5.57} \\
            \bottomrule
        \end{tabular}
    }
    \caption{\textbf{Experiments on different enhancement strategies.} We compare the effectiveness of atomic capability training (Cambrian-S-7B v.s. Cambrian-S-7B-Base), Chain-of-Thought (CoT) prompting, and our MSR-9K instruction tuning. The row $\Delta$ indicates the absolute performance gain achieved by our method compared to the base model.}
    \label{tab:cot_msr_results}
\end{table*}

\begin{figure}[t]
    \begin{center}
        \includegraphics[width=1\linewidth]{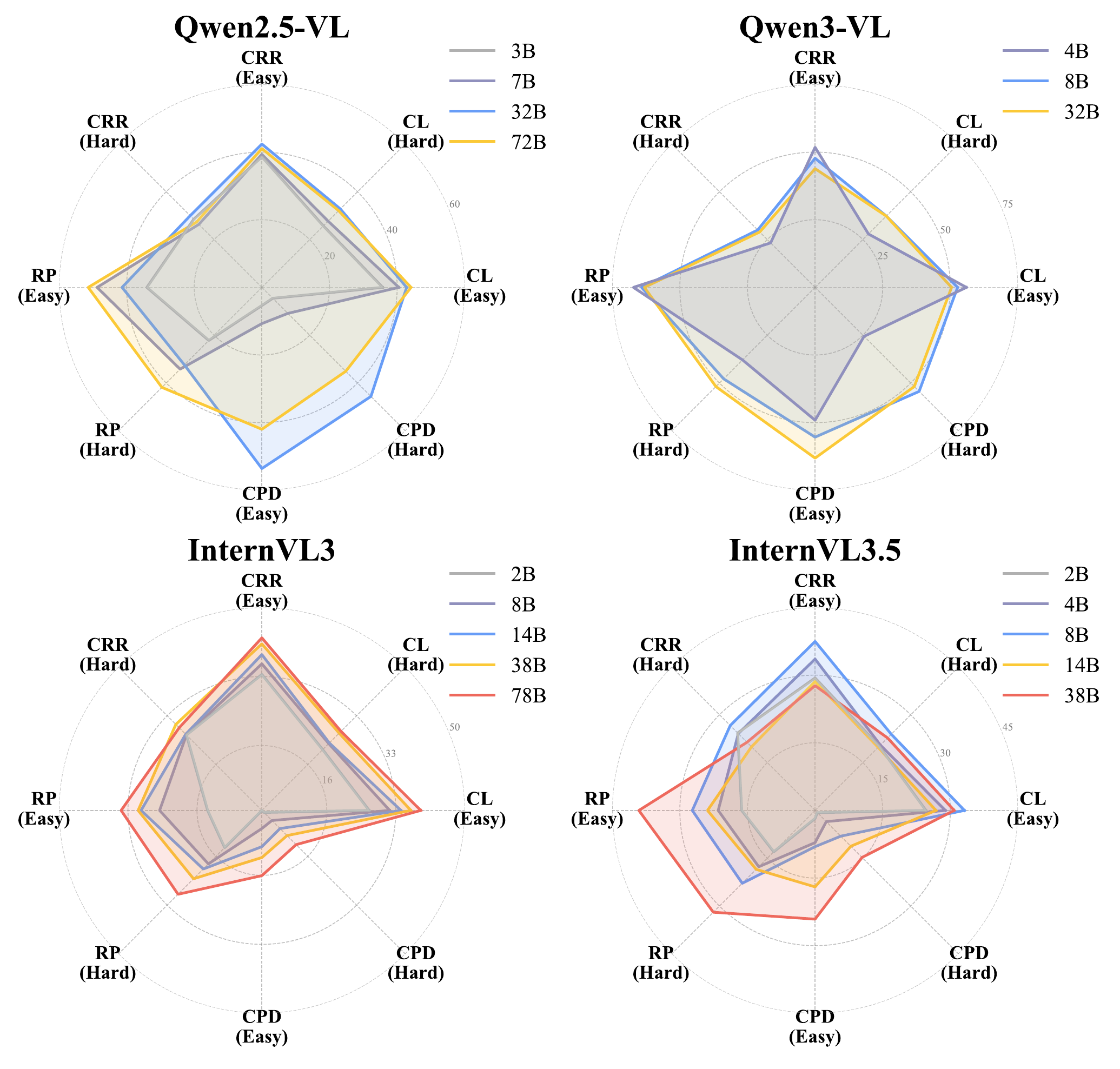}
    \end{center}
    \caption{The relationship between model size and MSR performance across various model families.}
    \label{fig:size_radar}
\end{figure}
\section{Experiments}

\subsection{Experimental Setup}

We evaluate a comprehensive set of 20 state-of-the-art open-source and proprietary models on Video-MSR, including Qwen3-VL~\cite{Qwen3-VL}, Qwen2.5-VL~\cite{Qwen2.5-VL}, InternVL-3.5~\cite{wang2025internvl35}, InternVL-3~\cite{zhu2025internvl3}, LLaVA-Next~\cite{liu2024llavanext}, Gemini-2.5~\cite{gemini25}, and GPT-4o~\cite{gpt4o}. To ensure a fair comparison, all open-source models are evaluated using 8$\times$NVIDIA A800 (80GB) GPUs, while proprietary models are accessed via their official APIs. 

Regarding the evaluation protocol, tasks are categorized into two formats. For Constrained Localization (CL) and Counterfactual Physical Deduction (CPD), which are formulated as multiple-choice questions, we employ Qwen2.5-7B-Instruct as a parser to extract the final predicted option and perform a direct string comparison against the ground truth. Conversely, for the open-ended tasks--Chain-based Reference Retrieval (CRR) and Route Planning (RP)--we utilize Gemini-3.0-Pro as an expert evaluator to assess the correctness and logical coherence of the model's responses. Across all experiments, we adopt a greedy decoding strategy for MLLMs to minimize randomness and ensure the reproducibility of our results. Further implementation details are provided in the supplementary material.

\subsection{Main Results}

We report the quantitative performance of various MLLMs on the Video-MSR benchmark. Table \ref{tab:main_results} summarizes the evaluation results across four distinct spatial reasoning tasks. In general, the Qwen3-VL series emerges as the most capable family among open-source models, while GPT-4o stands out as the strongest model among proprietary models, achieving overall accuracies of 43.92\% and 41.87\%, respectively.

\paragraph{Constrained Localization (CL).} In the CL task, Qwen3-VL-8B-Instruct achieves the highest accuracy (38.44\%), followed closely by GPT-4o (38.07\%). Notably, Gemini-2.5-Flash, despite being one of the most advanced proprietary models, significantly underperforms in this category. Its performance lags behind all other evaluated models, including the compact open-source InternVL3-2B-Instruct (29.73\%). This substantial gap highlights a critical deficiency in Gemini-2.5-Flash's ability to handle multi-constraint spatial filtering within complex, multi-object environments.

\paragraph{Chain-based Reference Retrieval (CRR).} Conversely, for CRR, Gemini-2.5-Flash secures the top spot with a score of 66.33\%, while Qwen3-VL-32B-Instruct achieves the second-highest score (57.33\%). It is worth noting that while both CRR and CL assess the ability to handle spatial dependencies, their cognitive demands differ: CRR emphasizes constructing a sequential retrieval chain to locate a final target, whereas CL necessitates multi-step filtering over a set of objects based on spatial attributes. The performance divergence suggests that while Gemini-2.5-Flash excels at sequential, single-threaded spatial localization, it struggles with the simultaneous processing of multi-object spatial information required by CL.

\paragraph{Route Planning (RP).} RP evaluates the active construction of spatial trajectories, sharing similar competency requirements with CRR. Consequently, as expected, Gemini-2.5-Flash and Qwen3-VL-32B-Instruct maintain their dominance, achieving 61.21\% and 57.93\% accuracy, respectively. In contrast, the InternVL series exhibits poor performance on RP, with accuracies hovering around 10\% or even approaching zero. Furthermore, LLaVA-NeXT-Video-7B fails to perform effective open-ended route planning, largely due to limitations in its open-ended generation capabilities.

\paragraph{Counterfactual Physical Deduction (CPD).} Finally, for CPD, which probes the model's capacity for multi-step spatial imagination based on visual cues, the Qwen3-VL-8B-Instruct and Qwen3-VL-32B-Instruct models achieve the highest scores (38.36\% and 36.53\%, respectively). Surprisingly, Gemini-2.5-Flash (24.77\%) and GPT-4o (25.57\%) perform poorly, ranking among the lowest alongside InternVL3.5-4B-Instruct (24.20\%). This result indicates a distinct gap between these proprietary systems and leading open-source models in performing implicit spatial reasoning and mental simulation.

\subsection{Further Analysis}

\paragraph{Larger models do not necessarily possess stronger MSR capabilities.} While the scaling law typically dictates that performance improves with parameter count, our analysis reveals a non-monotonic relationship between model size and MSR proficiency. As illustrated in Figure \ref{fig:size_radar}, the InternVL3 family generally adheres to the expected trend, where performance improves consistently with model size. However, the InternVL3.5 series exhibits significant volatility: while larger models demonstrate superior spatial reasoning in specific tasks, they paradoxically lag behind their smaller counterparts in others. For instance, InternVL3.5-38B-Instruct significantly outperforms smaller variants (2B--14B) on RP and CPD tasks, yet it yields the lowest performance on the \textit{Easy} subset of CRR. Furthermore, within the Qwen3-VL family, the compact 4B model unexpectedly secures the highest scores across the \textit{Easy} subsets of CL, CRR, and RP tasks. This discrepancy suggests that multi-hop spatial reasoning is not an emergent property solely dependent on scale. Instead, it likely requires specific architectural biases or high-quality data distributions that foster logical spatial connectivity, which cannot be compensated for by simply scaling up parameters trained on general-purpose multimodal data.

\begin{figure}[t]
    \begin{center}
        \includegraphics[width=\linewidth]{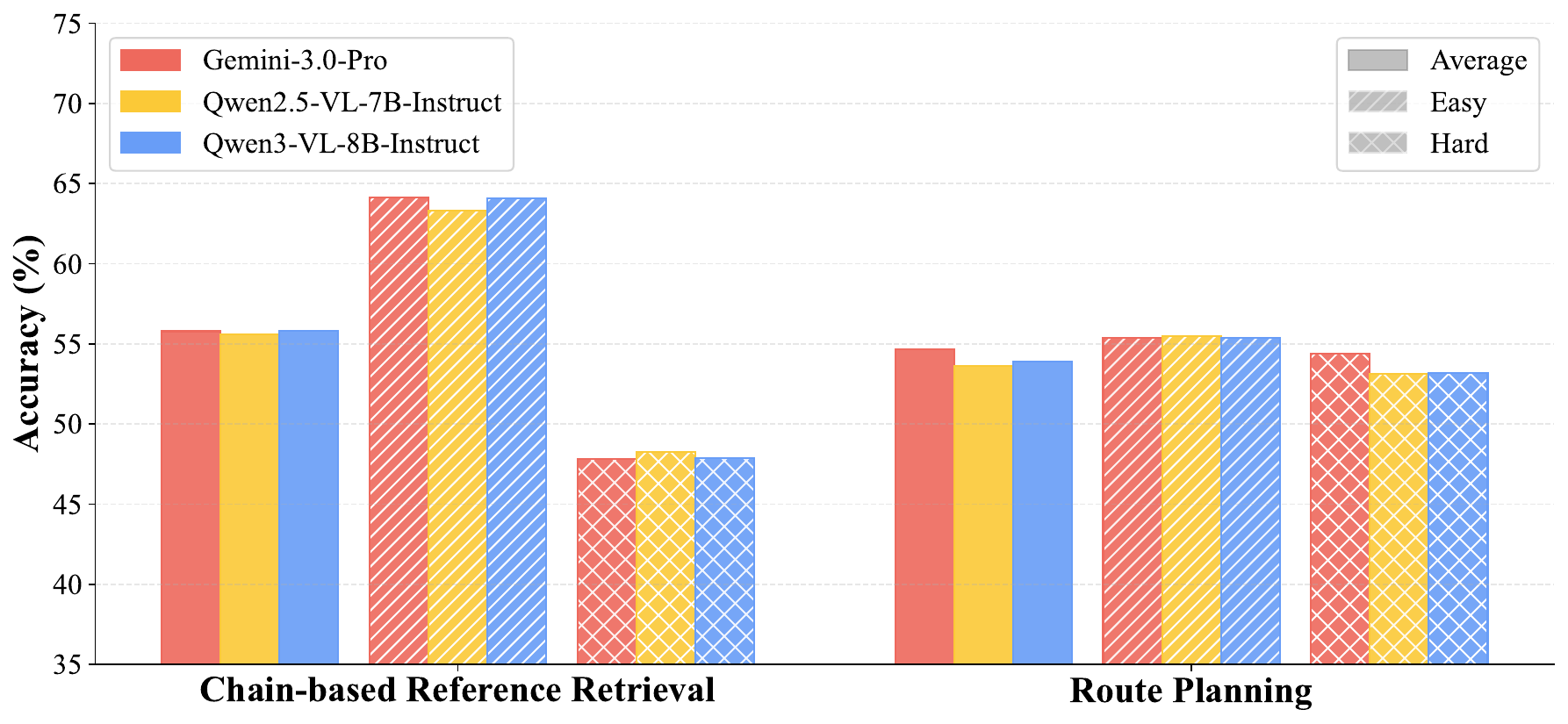}
    \end{center}
    \vspace{-0.5em}
    \caption{\textbf{Ablation study on the evaluator.} We employ three distinct models as evaluators for Video-MSR, ranging from Gemini-3.0-Pro to leading open-source models like Qwen-VL.}
    \vspace{-0.5em}
    \label{fig:evaluator_ablation}
\end{figure}

\paragraph{Chain-of-Thought (CoT) and atomic capability training fail to yield generalized improvements.} We observe that standard Chain-of-Thought (CoT) prompting, despite its proven efficacy in domains such as mathematics and coding, often proves detrimental in the Video-MSR context. As shown in Table \ref{tab:cot_msr_results}, applying CoT to Qwen2.5-VL-7B and Qwen3-VL-8B results in performance drops of 1.80\% and 4.54\%, respectively. This counter-intuitive finding is likely attributable to hallucinations and ``reasoning drift,'' where models fabricate non-existent spatial details during intermediate steps, leading to error propagation in the final deduction. Furthermore, models optimized for atomic visual reasoning capabilities struggle significantly with complex multi-hop spatial tasks. For instance, Cambrian-S-7B is fine-tuned on VSI-590K to master basic spatial concepts, such as size, distance, and perspective. However, compared to its base version, it surprisingly suffers a performance drop of over 5\% on Video-MSR (Table \ref{tab:cot_msr_results}). This indicates that strong fundamental perception alone does not automatically translate into the capacity to construct long-horizon spatial reasoning chains.

\paragraph{MSR-9K is significant for further improving the model's MSR capabilities.} To validate the effectiveness of our data construction paradigm, we fine-tuned models on MSR-9K. The results in Table \ref{tab:cot_msr_results} demonstrate substantial improvements: Qwen2.5-VL-7B-Instruct achieves a massive +7.82\% gain in overall accuracy, with the RP task witnessing a remarkable +48.62\% surge. Similarly, the already strong Qwen3-VL-8B-Instruct sees a consistent improvement of +3.10\%. These gains are pervasive across both ``Easy'' and ``Hard'' subsets. Notably, the improvement on the ``Hard'' subset for Qwen2.5-VL-7B-Instruct (+10.13\%) significantly exceeds that of the ``Easy'' subset (+4.79\%), confirming that MSR-9K does not merely teach the model to overfit to specific templates. Instead, it effectively bridges the domain gap, particularly in tasks like RP where base models lack priors, enhancing the model's ability to internalize spatial dependencies and plan trajectories.

\paragraph{Video-MSR is robust for evaluating models.} We implement a hybrid evaluation strategy tailored to different task formats. For multiple-choice tasks (CL and CPD), we assess performance via direct string matching between the parsed prediction and the ground truth. Conversely, for open-ended tasks (CRR and RP), we adopt an LLM-as-a-Judge paradigm, where the evaluator determines correctness by scrutinizing the model's response alongside the ground truth and video context. To ensure that our metric is objective and unbiased by the choice of the judge, we conducted an ablation study using three distinct evaluators: Gemini-3.0-Pro, Qwen2.5-VL-7B, and Qwen3-VL-8B. As depicted in Figure \ref{fig:evaluator_ablation}, the performance scores for a given target model remain highly consistent across all three evaluators, exhibiting negligible variance. This stability confirms that Video-MSR provides a rigorous evaluation benchmark, where performance gaps stem from the intrinsic difficulty of the spatial tasks rather than the subjectivity of the evaluator.

\section{Related Work}

\subsection{Video Understanding of MLLMs}
Video understanding, which necessitates the processing of abundant visual information across multiple frames, has emerged as a pivotal research domain alongside the rapid advancement of Multimodal Large Language Models (MLLMs)~\cite{Qwen-VL,internvl,llava}. Initial research efforts primarily focused on evaluating foundational perceptual capabilities on short clips~\cite{Activitynet-qa-short-clip,Next-qa-short-clip} and temporal awareness~\cite{liu-etal-2024-tempcompass}. Progressing beyond basic perception, subsequent studies have investigated MLLM capabilities in more complex scenarios, such as video-based mathematical reasoning~\cite{rasheed2025videomathqa} and long-form video comprehension~\cite{egoschema-long-video,wu2024longvideobench}. Furthermore, recent works have broadened the evaluation scope to encompass diverse reasoning tasks across various domains~\cite{videomme,li2024mvbench,shen2025fingercap}. Collectively, these benchmarks provide a systematic assessment of the video understanding capabilities of MLLMs.

\subsection{Benchmarks for Spatial Reasoning}
Building on the robust visual perception and semantic reasoning capabilities of MLLMs, evaluating their spatial reasoning has recently emerged as a core research focus. Consequently, numerous related benchmarks have been introduced. For example, VSR~\cite{vsr} comprehensively assesses the ability of MLLMs to recognize diverse spatial relationships. CV-Bench~\cite{cambrian-1-cvbench} and VSI-Bench~\cite{vsibench} broaden the evaluation scope by testing atomic spatial skills, including size comparison, orientation, counting, distance estimation, and temporal ordering. More recently, MMPerspective~\cite{tang2025mmperspective} has further advanced this field by specifically evaluating spatial reasoning across varying visual perspectives. However, these benchmarks remain limited to simplistic tasks relying on basic perception and reasoning. Consequently, how to systematically evaluate MLLMs' capabilities for complex spatial reasoning under strict constraints still remains underexplored.

\section{Conclusion}

In this work, we introduce Video-MSR, the first benchmark dedicated to assessing multi-hop spatial reasoning in video environments. Through a comprehensive evaluation of numerous MLLMs, we reveal a critical gap in current spatial intelligence: while models excel at visual perception, they struggle significantly with complex spatial logic, often susceptible to incoherence and hallucination during reasoning. Our analysis further uncovers that MSR proficiency is not strictly correlated with model scale, nor can it be trivially addressed by standard Chain-of-Thought prompting or atomic visual training. To bridge this gap, we curate MSR-9K, a high-quality instruction-tuning dataset. Fine-tuning on MSR-9K yields substantial performance gains, validating the necessity of logically grounded spatial data. We hope Video-MSR will serve as a pivotal foundation for future research, advancing the community toward developing MLLMs with stronger spatial reasoning capabilities.

\appendix

\section*{Ethical Statement}

We are committed to the ethical development of AI. The videos are sourced exclusively from publicly available datasets with permissible licenses, and we have ensured that no personally identifiable information is exposed. For the human verification process, all annotators were recruited with informed consent and compensated fairly. Furthermore, we have implemented strict filtering to ensure the dataset is free from social bias, toxicity, and harmful content.

\bibliographystyle{named}
\bibliography{ijcai26}

\end{document}